# MOTION TRANSFER-DRIVEN INTRA-CLASS DATA AUGMENTATION FOR FINGER VEIN RECOGNITION

*Xiu-Feng Huang[1], Lai-Man Po[1], Wei-Feng Ou[2]*

[1]City University of Hong Kong, [2]Dongguan University of Technology

**ABSTRACT**

Finger vein recognition (FVR) has emerged as a secure biometric technique because of the confidentiality of vascular bio-information. Recently, deep learning-based FVR has gained increased popularity and achieved promising performance. However, the limited size of public vein datasets has caused overfitting issues and greatly limits the recognition performance. Although traditional data augmentation can partially alleviate this data shortage issue, it cannot capture the real finger posture variations due to the rigid label-preserving image transformations, bringing limited performance improvement. To address this issue, we propose a novel motion transfer (MT) model for finger vein image data augmentation via modeling the actual finger posture and rotational movements. The proposed model first utilizes a key point detector to extract the key point and pose map of the source and drive finger vein images. We then utilize a dense motion module to estimate the motion optical flow, which is fed to an image generation module for generating the image with the target pose. Experiments conducted on three public finger vein databases demonstrate that the proposed motion transfer model can effectively improve recognition accuracy. Code is available at: https://github.com/kevinhuangxf/FingerVeinRecognition

## 1. INTRODUCTION

Finger vein recognition has emerged as a promising technology for secure identity verification. This is due to the unique characteristics of finger vascular patterns and the system's robustness against environmental factors such as varying lighting conditions. Extensive research and development efforts have resulted in significant advancements in FVR technologies [1,2].

Traditional FVR methods employ image processing techniques to extract a binary mask depicting the finger vein pattern for feature representation. Some well-known techniques include maximum curvature (MC) [3], wide line detector (WLD) [4], and repeated line tracking (RLT) [5]. On the other hand, deep learning-based approaches can utilize both vascular pattern extraction and feature embedding methods. For instance, FV-GAN [6] employs generative adversarial networks to segment vein patterns for recognition, while FusionAug [7] trains convolutional neural networks with fusion loss for feature embedding.

Recently, data-centric approaches, such as GAN-based methods [8,9], focus on generating synthetic finger vein datasets by designing random vein patterns and employing CycleGAN [10] to transform these patterns into realistic finger vein images. These techniques attempt to simulate finger motion by introducing random offsets to vein pattern coordinates. However, they have limitations in accurately modeling complex real-world motion. Traditional data augmentation techniques like rotation, cropping, and perspective transformation also struggle to replicate finger translations and rolling motions effectively. Pose transfer techniques have the capability to generate novel poses based on key point annotations [11,12]. Moreover, image animation methods like FOMM [13] and MRAA [14] can render input images in new poses by leveraging driving frames, all in a self-supervised manner. However, these approaches typically require additional annotations or videos as inputs, which pose challenges for online augmentation.

In this work, we propose a novel MT model that facilitates the generation of finger vein images in various poses while ensuring consistent vein patterns and finger geometry, which are crucial biometric features for accurate FVR. Our proposed MT model is lightweight and can be effectively applied for online intra-class augmentation by animating finger vein images to new poses. Furthermore, our approach complements synthetic finger vein image generation, and the combination of both techniques yields substantial improvements in FVR accuracy.

## 2. PROPOSED METHOD

In this section, we introduce our proposed method in detail. As introduced in Figure 1, our MT model can be divided into three parts: Key point and Pose Detection, Dense Motion Flow Estimation and Warping-based Image Generation. Given a source image S and a driving image D from the same subject's intra-class samples, the MT model will learn the motion between S and D to reconstruct the source image into a new image D′ to mimic the driving image's pose.

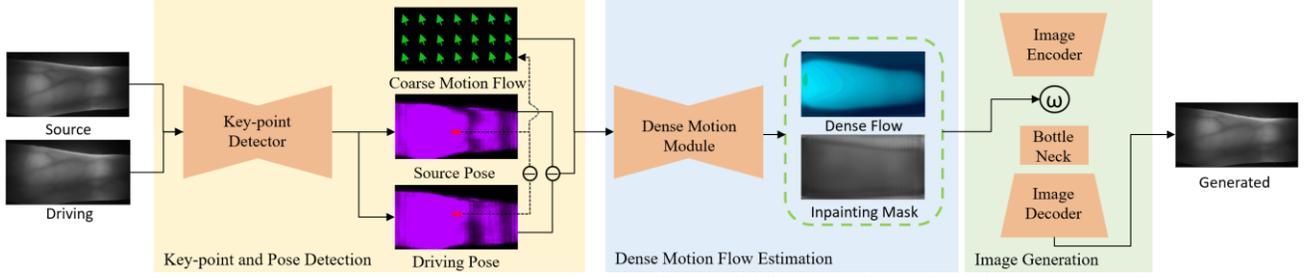

Figure 1. Overview of our proposed motion transfer model for FVR intra-class data augmentation

### 2.1 Key point and Pose Detection

Similar to the facial landmarks detection, we use a UNet architecture [15] network to extract the feature embedding and estimate the key point for the input image since its ability for localization [16]. We treat the finger as a unified entity instead of multiple articulated regions such as human poses. Let $x \in R^{h \times w \times 2}$ be the 2D pixel location in the input gray-scale image with height $h$ and width $w$, we activate the extracted feature output with the sigmoid function, so the key point location for the finger vein image can be estimated by the weighted sum of:

$$p = \sum_x \theta(x)\, x \qquad (1)$$

where $p$ is the pixel location of the detected key point, $\theta$ is the feature extracted from the UNet key point detector and activated by the sigmoid function. Based on the detected key point, we also estimate the key point covariance matrix $\Sigma$, which can indicate its orientation:

$$\Sigma = \sum_x \theta(x)(x-p)(x-p)^T \qquad (2)$$

The motivation for computing the key point and covariance is not only to capture the key point location and orientation, but also to build a detection confidence map by using the 2D Gaussian distribution, which indicates the heatmap $H$ of the finger pose:

$$H = \exp\left(-\frac{1}{2}(x-p)^T \Sigma^{-1}(x-p)\right) \qquad (3)$$

Thus, we obtain the key points $p_S$, $p_D$ and heatmaps $H_S$, $H_D$ for the source and driving images.

### 2.2 Dense Motion Flow Estimation

In this stage, we compute the key point locations difference $\Delta p = p_S - p_D$ the and the heatmap difference $\Delta H = H_S - H_D$. The key point locations difference $\Delta p$ can be seen as a sparse motion flow, and we repeat the key points location difference by $h \times w$ times as a coarse estimation of the dense motion flow. We use this coarse motion flow to warp the source image and concatenate the deformed source image with the heatmap difference $\Delta H$, which contains the appearance difference information and will become the input data for the Dense Motion Network. We still use a UNet architecture as the Dense Motion Network to learn a segmentation mask to indicate the significant area of the coarse motion flow. The final dense motion flow $\mathcal{F}$ will be estimated by the learned segmentation on coarse motion flow:

$$\mathcal{F} = M_{flow} \odot \left(f_r(\Delta p) \oplus \Delta H\right) \qquad (4)$$

where $M_{flow}$ is the learned segmentation mask for estimating dense motion flow, $f_r$ is the repeating operator, $\odot$ denotes the Hadamard product and $\oplus$ denotes the concatenation. The Dense Motion Network will also learn another mask $M_{inp}$ to indicate the inpainting area for the image generation.

### 2.3 Warping-based Image Generation

For the image reconstruction, we use the UNet architecture network with the flow warping strategy similar to FOMM. In specific, we have three down-sampling convolutional blocks for the encoder layers and three up-sampling convolutional blocks for the decoder layers. Before the $i^{th}$ decoder layer, the $i^{th}$ encoder layer extracts the source image to get the feature embedding $\xi_i$, the dense motion flow $\mathcal{F}$ is used to warp the feature embedding $\xi_i$, while the inpainting mask $M_{inp}$ is used to indicate the region that needs to be inpainted. The final transformed feature embedding $\xi_i'$ is written as:

$$\xi_i' = M_{inp} \odot f_\omega(\xi_i, \mathcal{F}) \qquad (5)$$

where $f_\omega$ is the warping operation. The inpainting mask $M_{inp}$ and dense motion flow $\mathcal{F}$ will be down-sampled to the corresponding size. The transformed feature embedding $\xi_i'$ will be passed to the subsequent decoder layer for the reconstruction of the target finger vein image.

### 2.4 Training Loss

Similar to FOMM, we use the perceptual loss to compare the reconstructed image $D'$ and the driving image $D$:

$$\mathcal{L}_{perc}(D', D) = \sum_{i}^{n} |N_i(D') - N_i(D)| \qquad (6)$$

where $N_i$ is the $i^{th}$ layer of the VGG19 pretrained network. We also use the equivariance loss for training the self-supervised key points detection:

$$\mathcal{L}_{eq} = |p_D - \Delta\hat{p}\hat{p}_D| \qquad (7)$$

where $\Delta\hat{p}$ is some random geometric transformation for the key point, and $\hat{p}_D$ is the transformed key point by $\Delta\hat{p}$. The final loss will sum these two losses: $\mathcal{L} = \mathcal{L}_{perc} + \mathcal{L}_{eq}$.

## 2.5 Motion Transfer-Driven Intra-class Augmentation

Following the training of our motion transfer model, we have the capability to utilize it as a motion transfer-driven intra-class augmentation (MT-Aug) for enriching the finger vein training samples. Prior to the training, we use the pretrained MT model to analyze the principal motion vectors on the training samples $X = [x_1, x_2, x_3, \ldots, x_N]$. We first use the key point detector $\theta$ to compute the key point difference $\Delta p$ between every two consecutive intra-class finger vein samples and collect into $\Delta P = [\Delta p_1, \Delta p_2, \ldots, \Delta p_m]$. We then use the PCA method on $\Delta P$ to analyze $n$ principal motion vectors $V = [v_1, v_2, \ldots, v_n]$ and set $n = 10$ which is practical enough. These learned principal motion vectors represent the major motion in the training samples, which can be used to spontaneously animate the finger vein images. In order to achieve a random augmentation effect, we finally use a linear combination method to generate a random motion vector $\hat{v}$:

$$\hat{v} = \sum_{i}^{n} a_i v_i \qquad (8)$$

where $a_i$ is a random generated scale factor with $\sum_{i}^{n} a_i = 1$ and $v_i$ is the $i^{th}$ principal vector in V.

During the FVR training, we will use the pretrained MT model to detect a source key point $p_S$ for a training sample $x$. We then generate a random motion vector $\hat{v}$ so that we can get a pseudo driving key point $\hat{p}_D = p_S + \hat{v}$. Based on the Eq. (2) and (3), we can compute the correspondence pseudo driving heatmap $\hat{H}_D$ and thus compute the heatmap difference $\Delta\hat{H} = H_S - \hat{H}_D$. The dense motion module M will use these $\hat{v}$ and $\Delta\hat{H}$ to generate the dense motion flow $\mathcal{F}$ and then the image generator $\xi$ will reconstruct a novel pose finger vein image. We summarize our proposed MT-Aug method into the Algorithm 1.

## 3. EXPERIMENTS

In this section, we conduct extensive experiments on three publicly available finger vein databases, validating the effectiveness of our proposed method.

---

**Algorithm 1 MT-Driven Intra-class Augmentation**

**Input:** Training samples X
**MT Model:** Key Point Detector $\theta$
  Dense Motion Module M
  Warping-based Image generator $\xi$
**Output:** Synthetic samples $x'$ with novel pose
a: Use $\theta$ to extract $\Delta p$ between intra-class samples in X and collect into $\Delta P = [\Delta p_1, \Delta p_2, \ldots, \Delta p_m]$
b: Use PCA method to analyze principal motion vectors $V = [v_1, v_2, \ldots, v_n]$ based on $\Delta P$.
During the FVR Training:
1: Detect the source key point $p_S$ for a training sample $x$ from X
2: Generate a random vector $v'$ based on Eq. (8)
3: Compute the pseudo driving key point $\hat{p}_D = p_S + \hat{v}$
4: Compute the pseudo driving heatmap $\hat{H}_D$ based on Eq. (2), (3) and the difference $\Delta\hat{H} = H_S - \hat{H}_D$.
5: Input $v'$ and $\Delta\hat{H}$ to M to estimate the motion flow $\mathcal{F}$
6: Use $\xi$ to reconstruct the $x'$ based on $\mathcal{F}$
7: Feed $x'$ to the FVR Baseline

### 3.1 Finger Vein Databases

We utilized the FV-USM[17], the MMCBNU[18] and the SDUMLA[19] databases. The FV-USM database contains 5904 finger vein images in total with 492 classes and 12 samples for each class captured from 123 subjects. The MMCBNU database contains 6000 finger vein images in total with 600 classes and 10 samples per class captured from 100 subjects. The SDUMLA database contains 3816 finger vein images in total with 628 classes and 6 samples per class captured from 106 subjects. We divided each database equally into training and testing sets. All finger vein images were preprocessed using the WLD method [4] to yield horizontally aligned ROI images sized 64×144 pixels.

### 3.2 Training Details

As illustrated in the Algorithm 1, during the FVR training, the MT model is applied as an augmentation step to enrich the dataset with novel pose finger vein samples. Our MT model remains lightweight, with a model parameter count of 233.7k and 0.1246 GFLOPs for a standard 64x144 aligned finger vein image. In terms of inference time, the model achieves a swift 10.68 ms on a i9-13900K Intel CPU. This efficiency empowers our MT model to function effectively as an online intra-class augmentation method.

We utilized the FusionAug [7] approach as the supervised baseline for training our FVR model. This method employs ResNet-18 [20] as the backbone network for feature extraction, employing a fusion loss combining cosine distance loss and triplet loss, along with a comprehensive intra-class augmentation pipeline including random resized cropping, rotation, perspective distortion, and color jittering.

The training employs a Stochastic Gradient Descent (SGD) optimizer and a batch size of 32. The initial learning rate is set to 0.01, and then decayed by a factor of 0.1 at epoch 40 and completed in 80 epochs. During testing, we calculate the cosine similarity between feature vectors for matching pairs to determine verification. Our performance evaluation utilizes the equal error rate (EER) as the performance evaluation metric.

Table 1. Performance comparison of different approaches for biometric vein verification.

| Methods | FVUSM | MMCBNU | SDUMLA |
|---|---|---|---|
| WLD | 5.43% | 3.70% | 3.83% |
| RLT | 4.67% | 1.68% | 5.88% |
| MC | 2.63% | 1.12% | 3.67% |
| FusionAug | 0.45% | 0.30% | 3.31% |
| FusionAug+MT-Aug | 0.35% | 0.19% | 2.91% |

Table 2. Performance of training with synthetic finger vein samples.

| Database | Inter-class | Intra-class | EER |
|---|---|---|---|
| FVUSM | 246 synthetic samples | w/o MT-Aug | 0.41% |
| | | with MT-Aug | 0.19% |
| MMCBNU | 300 synthetic samples | w/o MT-Aug | 0.26% |
| | | with MT-Aug | 0.16% |
| SDUMLA | 318 synthetic samples | w/o MT-Aug | 3.19% |
| | | with MT-Aug | 2.46% |

Synthetic finger vein generation[9, 21] often struggles to faithfully replicate the motion observed among intra-class samples within real datasets. To address this limitation, we extend our MT model to incorporate synthetic finger vein samples. Similar to Style-VIG[22], we employ the StyleGAN2[23] method to train on the finger vein datasets. The synthetic finger vein samples have enough variation for vein pattern and finger geometry so we treat each sample as a distinct inter-class sample. We use the StyleGAN2 to generate synthetic samples of the same amount with their correspondence dataset inter-classes to double the total classes. For each synthetic finger vein sample, we utilized the proposed MT model to generate novel poses similar to the real dataset based on the learned motion representation.

**3.3 Performance Evaluation and Analysis**

In this section, we evaluate the efficacy of our proposed MT model for augmenting finger vein intra-class samples. As depicted in the Table 1, the experimental results show obvious enhancement achieved by our proposed MT-Aug method across all the three public datasets, compared with the baseline outcomes.

When incorporating with the synthetic samples, as shown in the Table 2, we observe that while the increasement of the synthetic finger vein images as the inter-class samples can improve the baseline model performance, the improvement could be minor and even smaller than the enhancement obtained using our proposed MT-Aug without synthetic samples, as detailed in Table 1. It's worth noting that our proposed MT-Aug method can make a significant improvement compared with the baseline result, especially for the SDUMLA database. This database is more challenging due to its higher motion complexity when compared to the FVUSM and MMCBNU databases. These results show that our proposed MT model serves an effective intra-class augmentation method, especially in the context of enhancing the performance when employing synthetic finger vein samples.

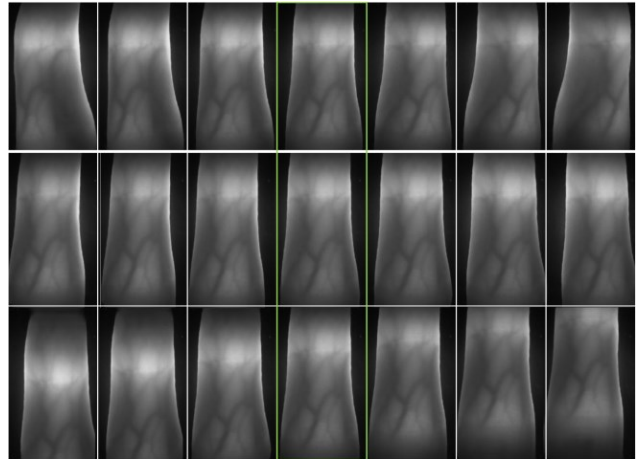

Figure 2. Visualization for using our proposed method to achieve finger rolling, horizontal and vertical translation. The fourth (middle) column is the testing image from our in-house dataset.

We make visualizations of our motion transfer results, using our in-house dataset, which covers a variety of finger geometric variations such as translation, rotation and rolling. Our in-house dataset contains 252 videos for different fingers and each video contains 120 continuous finger vein images. We still split the dataset equally into training set and testing set. By training our proposed MT model on the in-house dataset, we can learn generic motion representation to make an obvious pose transformation effect. As shown in the Figure 2, employing the learned principal motion vectors can facilitate seamless animation of finger rolling and translation for the finger vein image in the testing set. This outcome proves our proposed method's proficiency in generating finger vein images with novel poses for the intra-class augmentation purpose.

**4. CONCLUSION**

In this paper, we present a novel framework for motion transfer-driven intra-class augmentation of finger vein images that enables the generation of realistic variations in finger motion without the need for external pose or image signals. By leveraging our method, we can generate high-fidelity synthetic images with intra-class samples that closely resemble real finger vein datasets, and obtain promising recognition performance. Our framework facilitates privacy protection by training deep neural networks utilizing completely synthetic finger vein samples instead of real biometric data.

# 5. REFERENCES

[1] R. Das, E. Piciucco, E. Maiorana, and P. Campisi, "Convolutional Neural Network for Finger-Vein-Based Biometric Identification," *IEEE Transactions on Information Forensics and Security*, vol. 14, no. 2, pp. 360–373, Feb. 2018, doi: 10.1109/TIFS.2018.2850320.

[2] B. Hou, H. Zhang, and R. Yan, "Finger-Vein Biometric Recognition: A Review," *IEEE Trans Instrum Meas*, 2022, doi: 10.1109/TIM.2022.3200087.

[3] N. Miura, A. Nagasaka, and T. Miyatake, "Extraction of Finger-Vein Patterns Using Maximum Curvature Points in Image Profiles."

[4] B. Huang, Y. Dai, R. Li, D. Tang, and W. Li, "Finger-vein authentication based on wide line detector and pattern normalization," in *Proceedings - International Conference on Pattern Recognition*, 2010, pp. 1269–1272. doi: 10.1109/ICPR.2010.316.

[5] N. Miura, A. Nagasaka, and T. Miyatake, "Feature extraction of finger-vein patterns based on repeated line tracking and its application to personal identification," *Mach Vis Appl*, vol. 15, no. 4, pp. 194–203, Oct. 2004, doi: 10.1007/s00138-004-0149-2.

[6] W. Yang, C. Hui, Z. Chen, J. H. Xue, and Q. Liao, "FV-GAN: Finger Vein Representation Using Generative Adversarial Networks," *IEEE Transactions on Information Forensics and Security*, vol. 14, no. 9, pp. 2512–2524, Sep. 2019, doi: 10.1109/TIFS.2019.2902819.

[7] W. F. Ou, L. M. Po, C. Zhou, Y. A. U. Rehman, P. F. Xian, and Y. J. Zhang, "Fusion loss and inter-class data augmentation for deep finger vein feature learning," *Expert Syst Appl*, vol. 171, Jun. 2021, doi: 10.1016/j.eswa.2021.114584.

[8] W. F. Ou, L. M. Po, C. Zhou, P. F. Xian, and J. J. Xiong, "GAN-Based Inter-Class Sample Generation for Contrastive Learning of Vein Image Representations," *IEEE Trans Biom Behav Identity Sci*, vol. 4, no. 2, pp. 249–262, Apr. 2022, doi: 10.1109/TBIOM.2022.3152345.

[9] H. Yang, P. Fang, and Z. Hao, "A GAN-based Method for Generating Finger Vein Dataset," in *ACM International Conference Proceeding Series*, Association for Computing Machinery, Dec. 2020. doi: 10.1145/3446132.3446150.

[10] J.-Y. Zhu, T. Park, P. Isola, and A. A. Efros, "Unpaired Image-to-Image Translation using Cycle-Consistent Adversarial Networks," Mar. 2017.

[11] Y. Men, Y. Mao, Y. Jiang, W.-Y. Ma, and Z. Lian, "Controllable Person Image Synthesis with Attribute-Decomposed GAN," Mar. 2020.

[12] Z. Cao, G. Hidalgo, T. Simon, S.-E. Wei, and Y. Sheikh, "OpenPose: Realtime Multi-Person 2D Pose Estimation using Part Affinity Fields," Dec. 2018.

[13] A. Siarohin, S. Lathuilière, S. Tulyakov, E. Ricci, and N. Sebe, "First Order Motion Model for Image Animation," Feb. 2020.

[14] A. Siarohin, O. J. Woodford, J. Ren, M. Chai, and S. Tulyakov, "Motion Representations for Articulated Animation," Apr. 2021.

[15] O. Ronneberger, P. Fischer, and T. Brox, "U-Net: Convolutional Networks for Biomedical Image Segmentation," May 2015.

[16] J. P. Robinson, Y. Li, N. Zhang, Y. Fu, and and S. Tulyakov, "Laplace Landmark Localization," Mar. 2019.

[17] M. S. Mohd Asaari, S. A. Suandi, and B. A. Rosdi, "Fusion of Band Limited Phase Only Correlation and Width Centroid Contour Distance for finger based biometrics," *Expert Syst Appl*, vol. 41, no. 7, pp. 3367–3382, Jun. 2014, doi: 10.1016/j.eswa.2013.11.033.

[18] Y. Lu, S. J. Xie, S. Yoon, Z. Wang, and D. S. Park, "An available database for the research of finger vein recognition," in *2013 6th International Congress on Image and Signal Processing (CISP)*, IEEE, Dec. 2013, pp. 410–415. doi: 10.1109/CISP.2013.6744030.

[19] Y. Yin, L. Liu, and X. Sun, "SDUMLA-HMT: A Multimodal Biometric Database," 2011, pp. 260–268. doi: 10.1007/978-3-642-25449-9_33.

[20] K. He, X. Zhang, S. Ren, and J. Sun, "Deep Residual Learning for Image Recognition," in *2016 IEEE Conference on Computer Vision and Pattern Recognition (CVPR)*, IEEE, Jun. 2016, pp. 770–778. doi: 10.1109/CVPR.2016.90.

[21] J. Zhang, Z. Lu, M. Li, and H. Wu, "GAN-Based Image Augmentation for Finger-Vein Biometric Recognition," *IEEE Access*, vol. 7, pp. 183118–183132, 2019, doi: 10.1109/ACCESS.2019.2960411.

[22] E. H. Salazar-Jurado, R. Hernández-García, K. Vilches-Ponce, R. J. Barrientos, M. Mora, and G. Jaswal, "Towards the generation of synthetic images of palm vein patterns: A review," *Information Fusion*, vol. 89, pp. 66–90, Jan. 2023, doi: 10.1016/j.inffus.2022.08.008.

[23] T. Karras, S. Laine, M. Aittala, J. Hellsten, J. Lehtinen, and T. Aila, "Analyzing and Improving the Image Quality of StyleGAN," Dec. 2019.